\documentclass[letterpaper, 10 pt, journal, twoside]{ieeeconf}

\usepackage[utf8]{inputenc}
\usepackage{amsmath,bm} 
\usepackage{amssymb}
\usepackage{graphicx}
\usepackage{blindtext}
\usepackage{wrapfig}
\usepackage{subcaption}
\usepackage{color,soul}
\usepackage{cite}
\usepackage{mathtools}
\usepackage{eufrak}
\usepackage[letterpaper, top=60pt,bottom=43pt,left=48pt,right=48pt]{geometry} 

\usepackage{encryptedcontrol}

\begin{document}

\title{
Perfectly Undetectable False Data Injection Attacks on Encrypted Bilateral Teleoperation System based on Dynamic Symmetry and Malleability
}

\author{Hyukbin Kwon$^{1}$, Hiroaki Kawase$^{2}$, Heriberto Andres Nieves-Vazquez$^{3}$ , Kiminaro Kogiso$^{2}$, and Jun Ueda$^{1}$%
\thanks{
This work was supposed in part by NSF CMMI Grant 2112793 and JSPS KAKENHI Grand Number JP22H01509 and JP23K22779. Hiroaki Kawase was also supported by JST SPRING, Grant Number JPMJSP2131.}\thanks{$^{1}$ Hyukbin Kwon and Jun Ueda are with the George W. Woodruff School of Mechanical Engineering, Georgia Institute of Technology, Atlanta, GA 30332-0405, USA. (e-mail: {\tt\footnotesize bin.kwon@gatech.edu, jun.ueda@me.gatech.edu}).}
\thanks{$^{2}$ Hiroaki Kawase and Kiminao Kogiso are with the Department of Mechanical and Intelligent Systems Engineering, The University of Electro-Communications, Chofu, Tokyo 1828585, Japan. (e-mail: {\tt\footnotesize kawase@uec.ac.jp, kogiso@uec.ac.jp}).}%
\thanks{$^{3}$ Heriberto Andres Nieves-Vazquez is with the Department of Biomedical Engineering, Georgia Institute of Technology, Atlanta, GA, 30332, USA. (e-mail: {\tt\footnotesize hnieves@gatech.edu}).} 
}

\maketitle

\begin{abstract}
This paper investigates the vulnerability of bilateral teleoperation systems to perfectly undetectable False Data Injection Attacks (FDIAs). Teleoperation, one of the major applications in robotics, involves a leader manipulator operated by a human and a follower manipulator at a remote site, connected via a communication channel. While this setup enables operation in challenging environments, it also introduces cybersecurity risks, particularly in the communication link. The paper focuses on a specific class of cyberattacks: perfectly undetectable FDIAs, where attackers alter signals without leaving detectable traces at all. Compared to previous research on linear and first-order nonlinear systems, this paper examines bilateral teleoperation systems with second-order nonlinear manipulator dynamics. The paper derives mathematical conditions based on Lie Group theory that enable such attacks, demonstrating how an attacker can modify the follower manipulator's motion while the operator perceives normal operation through the leader device. This vulnerability challenges conventional detection methods based on observable changes and highlights the need for advanced security measures in teleoperation systems. To validate the theoretical results, the paper presents experimental demonstrations using a teleoperation system connecting robots in the US and Japan.
\end{abstract}

{\fontsize{9}{12}\selectfont
}


\section{Introduction}
Teleoperation of a remote manipulator is one of the traditional and still important applications in robotics. While a variety of system configurations have been implemented, one specific configuration places a  manipulator at the remote site, such as space, a plant with high radiology, a magnetic field, or even inside of the human body, as a follower, and places a manipulator with a similar structure at the user’s site as the leader. A human operator physically operates the leader manipulator in a way similar to a joystick, and the recorded motion is transferred via a communication channel to the remote follower manipulator to be exactly reproduced.  When the motion of the follower manipulator, together with forces from interactions with the environment, is sent back to the leader manipulator, and vice versa, the architecture is called bilateral teleoperation\cite{dong2019false, li2011design, ueda2004force}; otherwise, it is called unilateral teleoperation.

When a robot teleoperation system is seen as a cyber-physical system (CPS) from the cybersecurity standpoint, the communication channel is the weakest component in terms of security, where an attacker would attempt to intercept, modify, abandon, or inject malicious signals to impact the system integrity and performance. Among possible cyberattack modes, false data injection attacks (FDIAs) literally alter signals on the communication lines based on the attacker’s certain knowledge about the system \cite{musleh2019survey, sargolzaei2019detection,pang2021detection}.

In many cyberattack cases, the attacker’s strategy can be determined in terms of Game Theory, where an attacker would minimize the risk of being detected while maximizing their impacts. This motivated the study of stealthy attacks where malicious changes in the system’s state are small and thus may not be detected by an attack detector that was implemented as a countermeasure by the user \cite{ashok2016online,li2019optimal, guo2022stealthy}. Detection of FDIAs may be conducted by observing subtle changes in state variables using a statistical method \cite{zhao2018generalized} and watermarking \cite{mo2009secure,miao2013stochastic}.

\begin{figure}[b]
    \centering
    \includegraphics[width = \columnwidth]{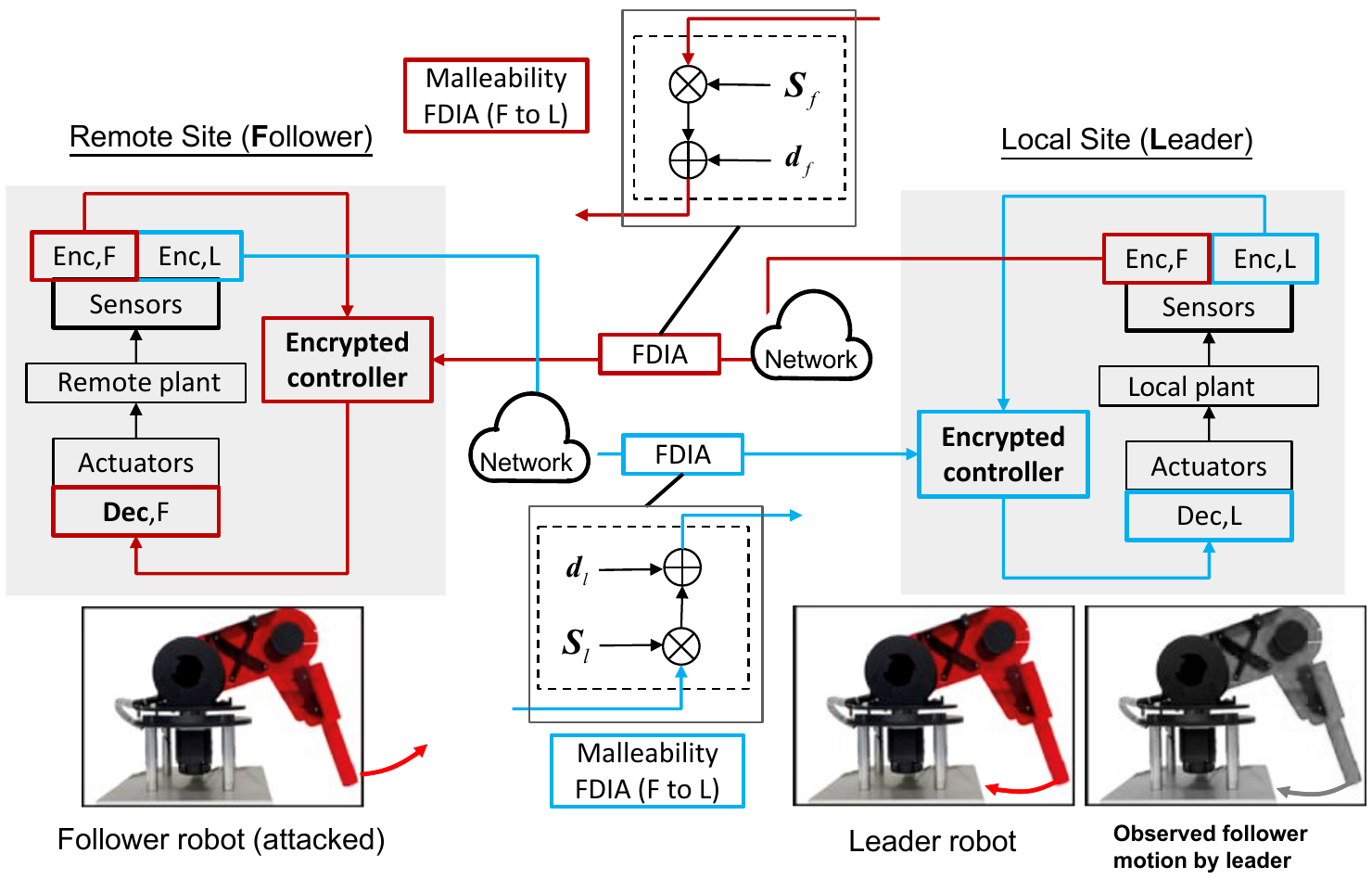}
    \caption{Conceptual diagram of malleability FDIA applied to a 4 channel encrypted bilateral teleoperation system.}
    \label{fig:concept}
\end{figure}

This paper considers an extreme case of stealthy attacks called perfectly undetectable attacks where an intelligent attacker successfully implements an FDIA so that ``no changes" in the signals sent back from the plant are observed by the user \cite{sandberg2022secure,weerakkody2016graph,cam2014modeling}. The authors have studied this specific type of FDIAs in a linear manipulator control system as well as a first-order nonlinear mobile robot control system\cite{ueda2024affinetransformationbasedperfectlyundetectable,ueda2024perfectlyundetectablereflectionscaling}. This paper will reveal that a class of bilateral teleoperation systems with typical second-order nonlinear manipulator dynamics may be susceptible to perfectly undetectable FDIAs where the user who manipulates the leader device perceives the operation as if it were normal, while the motion of the follower is indeed altered at the remote site.  Mathematical conditions to enable such perfectly undetectable FDIAs will be derived and experimentally demonstrated using a bilateral teleoperation system connecting two robots in the US and Japan. 

The primary objective of this paper is to identify conditions that achieve affine transformation-based perfectly undetectable FDIA \cite{ueda2024affinetransformationbasedperfectlyundetectable} on a bilateral teleoperation system with nonlinear dynamics as illustrated in Fig. \ref{fig:concept}. Compared to covert attacks that solve undetectable attacks as an optimization problem with full knowledge of the plant dynamic model, affine transformations can be implemented in a much simpler form, potentially impacting encrypted control systems by utilizing a security hole known as malleability. 

\section{Perfectly Undetectable FDIA}
\begin{figure}
    \centering\includegraphics[width=1.0\linewidth]{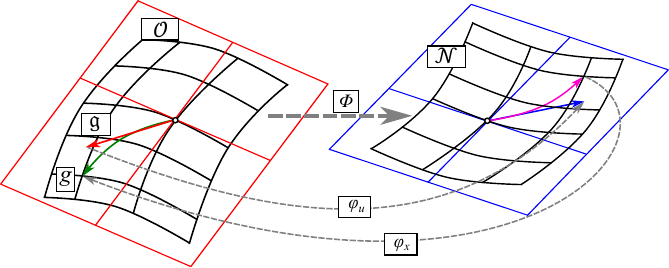}
    \caption{Visualization of Perfectly Undetectable FDIA Through Lie Group.}
    \label{fig:lieConcept}
\end{figure}

\subsection{Attackability analysis based on Lie Groups}

The FDIA is defined in the form of an affine transform both on the controller commands and observables.
A general control-affine nonlinear plant to be controlled remotely
with a control input $\bm{u} \in \mathbb{R}^m$ and observables $\bm{x} \in \mathbb{R}^n$ is given as:
\begin{equation}
    \label{controlAffinePlant}
    \dot {\bm  x} = \bm f(\bm x) + \bm h(\bm x)\bm u.
\end{equation}




FDIA may be systematically applied to (\ref{controlAffinePlant}) in the form of transformations on the control command $\bm{u}$ and observables $\bm{x}$ assuming full state feedback control.
This FDIA attack can be described in terms of maps $\varphi_x : \mathbb{R}^n \rightarrow \mathbb{R}^n$ and $\varphi_u : \mathbb{R}^m \rightarrow \mathbb{R}^m$.
Given plant dynamics, a perfectly undetectable FDIA is possible if there exist pairs of maps that satisfy conditions in Appendix \ref{appen_perfectFDIAplant}.

Lie Groups are a useful abstraction of continuous motion and can be used to describe plant dynamics\cite{LieOverview}.
Existence of perfectly undetectable FDIAs may be described as an automorphism $\varphi_u$ applied to the lie algebra $\mathfrak{g}$ of the original group $\bm{G}$, which then induces a global automorphism $\Phi$.
Since the inverse $\Phi^{-1}$ is always defined, it can be used to recover the original group $\bm{G}$:


\begin{align}
    \label{phiudef}
    \varphi_u &: \mathfrak{g} \rightarrow \mathfrak{g} \text{ is an automorphism}\\
    \exp{(\varphi_u (\mathfrak g))} &= \Phi(\bm g)\\
    \Phi(\bm x) &= \exp (\varphi_u^{-1} (\ln{(\bm x)})\\
    \label{phixdef}
    \varphi_x( \bm x) &= \Phi(\bm {x})^{-1} = \exp (-\varphi_u^{-1} (\ln{\bm x)}) 
\end{align}
The effect of these Lie Algebra elements are visualized on a 3D space in Fig. \ref{fig:lieConcept}.
$\bm{g}$, an element in the group $\bm{G}$ can be mapped to an element of $\mathcal{O}$ via the exponential map.
$\varphi_u$ and $\Phi$ maps the Lie algebra $\mathfrak g $and element of group $G$ to the manifold $N$ and the Lie group that spans it.

It is possible to formulate a Lie Group composite to describe more complex systems such as the manipulator used in this study.
A composite manifold, defined as $\hat {\mathcal{O}} = \langle \mathcal{O}_1, \cdots \mathcal{O}_o \rangle$ fulfills the group axioms.
If a candidate map $\hat \varphi_u$ is an automorphism for all Lie Groups defined over the each manifold in the composite, then a perfectly undetectable FDIA is possible on the entire system.
The $\hat \varphi_u$ for the composite case can be created from a block diagonal matrix of $\varphi_u$ for each manifold. This ensures that $\hat \varphi_u$ is also a valid automorphism.


\subsection{Perfectly undetectable FDIA on manipulator dynamics}
\label{perfectFDIAsinglemanipulator}

The paper adopts two identical 2-degree-of-freedom (DOF) robotic manipulators resembling the first two joints of PUMA manipulators where the first joint rotates about the vertical axis and the second rotates about a horizontal axis perpendicular to the first joint axis. As described in Section \ref{experiments}, one manipulator is placed in Atlanta, US, and the other in Tokyo, Japan.

A typical dynamic model of one of the robotic manipulators is considered:
\begin{align}
 (m_p l^2_2 \cos{\theta_2}^2 + J_1) \Ddot{\theta}_1 - 2 m_p l^2_2 \cos{\theta_2} \sin(\theta_2)\dot\theta_1 \dot\theta_2 = \tau_1 \label{robotdeom1}\\
m_p l_2^2 \ddot \theta_2 + m_p l_2^2\cos{\theta_2}\sin{\theta_2}\dot\theta_1^2+m_p g l_2\cos{\theta_2} = \tau_2
\label{robotdeom2}
\end{align}
\noindent where $\theta_1$ and $\theta_2$ are the yaw and pitch joint angles, $J_1$ is the moment of inertial of the first link (the component 3D printed in black) including the motor about $\theta_1$, and $g$ is the gravity constant. The inertia of Link 2 (the component 3D printed in red) is relatively low and practically dominated by a handle that is modeled as a point-mass $m_p$ for simplicity that is located at $l_2$ from the Joint 2 axis. $\tau_1$ and $\tau_2$ are motor torques of the yaw and pitch axes, respectively. 

In later sections, subscript ``$l$" is used for the leader manipulator, e.g., $\theta_{1l}$ for Joint 1 angle, and ``$f$" is used for the follower manipulator accordingly, e.g., $\tau_{2f}$. All geometric and dynamic parameters are identical between two manipulators. Subscripts are omitted when only a single manipulator is discussed. 

{\bf Proposition 1: Perfectly undetectable reflection FDIA on the yaw axis.} For the single manipulator dynamics (\ref{robotdeom1}) (\ref{robotdeom2}), only reflection about the yaw axis $\theta_1$ and $\tau_1$ achieves perfectly undetectable FDIA. 

{\it Sketch of proof:}
The symmetries of the manipulator dynamics can be identified as follows.
The configuration of a 2R manipulator can be parameterized into a bundle of two Lie groups in $S^1$ each representing joint angles $\theta_1$ and $\theta_2$.
In order to formulate a perfect FDIA, (\ref{phiudef}) and (\ref{phixdef}) should be satisfied.
To identify candidates for $\varphi_u$, automorphisms of the manipulator need to be identified through inspection of the nonlinear dynamic equations. Considering (\ref{robotdeom1}), the transform $[\pm1, \pm1]^T$ are automorphisms.
Disregarding the trivial attack, $[-1, \pm1]^T$ induces an inversion of the joint torques.
On the second manifold (\ref{robotdeom2}), the automorphisms can be identified to be $[\pm1,1]$. $\blacksquare$

Here the obvious choice for a successful attack is negation of $\theta_1$ which results in a reflection attack about its initial condition represented as $\tilde \theta_1=-\theta_1 + 2 \theta_1(0)$.
This gives the FDIA set $\hat \varphi_u = \text{diag}(-1,1)$ and $\hat \varphi_x = \text{diag}(-1,1)$.
Also, the corresponding joint effort is attacked as $\tilde \tau_1=-\tau_1$. While no attack is applied to the pitch axis, let's introduce $\tilde \theta_2=\theta_2$ and $\tilde \tau_2=\tau_2$ for simplicity. 
Substituting the attacked observables and commands into (\ref{robotdeom1}) and (\ref{robotdeom2}) yields:
\begin{align}
 (m_p l^2_2 \cos{\tilde \theta_2}^2 + J_1) \Ddot{\tilde \theta}_1 - 2 m_p l^2_2 \cos{\tilde \theta_2} \sin(\tilde \theta_2) \dot{\tilde{ \theta_1}} \dot {\tilde{ \theta_2}} = \tilde \tau_1 \label{robotdeom1reflected}\\
m_p l_2^2 \ddot{\tilde{ \theta_2}} + m_p l_2^2\cos{\tilde \theta_2}\sin{\theta_2}\dot{\tilde {\theta_1}}^2+m_p g l_2\cos{\tilde \theta_2} = \tilde \tau_2, 
\label{robotdeom2reflected}
\end{align}
that are identical to the nominal model, achieving perfectly undetectable FDIA from the controller's perspective {\it regardless} of the control scheme \cite{ueda2024affinetransformationbasedperfectlyundetectable}.

{\bf Remark 1:} If the gravity term is locally compensated, reflection about the pitch axis, $\tilde \theta_2=-\theta_2$ (note that this reflection must be about the horizontal position as its neutral angle), will also achieve perfectly undetectable FDIA. Otherwise, the opposite sign will appear in the gravity term in (\ref{robotdeom2}).

\subsection{Perfectly undetectable FDIA on 4-channel bilateral teleoperation system}

To represent the left-hand side terms of (\ref{robotdeom1}) and (\ref{robotdeom2}) in simpler forms, functions $f_{1l}, f_{2l}, f_{1f}, f_{2f}$ are introduced:

\noindent Leader:
\begin{align}
 f_{1l}(\dot{\theta}_{1l},\ddot{\theta}_{1l},\theta_{2l}, \dot{\theta}_{2l})= \tau_{1l} + \tau^e_{1l} 
 \label{robotdeom1lteleop}\\
 f_{2l}(\dot{\theta}_{1l},\theta_{2l}, \ddot{\theta}_{2l})= \tau_{2l} - \tau^e_{2l} 
\label{robotdeom2lteleop}
\end{align}

\noindent Follower:
\begin{align}
f_{1f}(\dot{\theta}_{1f},\ddot{\theta}_{1f},\theta_{2f}, \dot{\theta}_{2f})= \tau_{1f} + \tau^e_{1f}
\label{robotdeom1fteleop}\\
f_{2f}(\dot{\theta}{1f},\theta_{2f}, \ddot{\theta}_{2f})= \tau_{2f} - \tau^e_{2f}
\label{robotdeom2fteleop}
\end{align}

\noindent where $\tau^e_{1l}$ and $\tau^e_{2l}$ 
are equivalent moments to the external force applied by the user at the handle. $\tau^e_{1f}$ and $\tau^e_{2f}$ are equivalent moments to the external force from the environment at the follower's handle. 

The leader and follower exchange both displacements and external forces to perform force-reflecting bilateral control \cite{ueda2004force}. 
As illustrated in Fig. \ref{fig:concept}, the paper considers affine transformations using static parameters, $\bm{S}_l, \bm{d}_l, \bm{S}_f, \bm{d}_f$, as a possible implementation of FDIA, computing attacked signals.
In the communication channel from the follower to the leader, attacked signals are given as:
\begin{eqnarray}
\left[
\begin{array}{c}
\tilde{\theta}_{1f} \\
\tilde{\theta}_{2f} \\
\tilde{\tau}^e_{1f} \\
\tilde{\tau}^e_{2f}
\end{array}
\right] = \bm{S}_l 
\left[
\begin{array}{c}
\theta_{1f} \\
\theta_{2f} \\
\tau^e_{1f} \\
\tau^e_{2f}
\end{array}
\right] + \bm{d}_l.
\end{eqnarray}
\noindent Likewise, the attack to the communication channel from the leader to the follower yields:
\begin{eqnarray}
\left[
\begin{array}{c}
\tilde{\theta}_{1l} \\
\tilde{\theta}_{2l} \\
\tilde{\tau}^e_{1l} \\
\tilde{\tau}^e_{2l}
\end{array}
\right] = \bm{S}_f 
\left[
\begin{array}{c}
\theta_{1l} \\
\theta_{2l} \\
\tau^e_{1l} \\
\tau^e_{2l}
\end{array}
\right] + \bm{d}_f.
\end{eqnarray}

In normal operation (no attack), a bilateral controller is designed to realize$\theta_{1l,2l}=\theta_{1f,2f}$ and $\tau^e_{1l,2l}=\tau^e_{1f,2f}$ as the ideal response, with errors due to the intervening impedance.
Under FDIA, control commands, each using the signals from the counterpart joint, are determined as follows:
$\tau_{1l}(t)=\tau_{1l}(\theta_{1l}, \tilde{\theta}_{1f}, \tau^e_{1l}, \tilde{\tau}^e_{1f})$,
$\tau_{2l}(t)=\tau_{2l}(\theta_{2l}, \tilde{\theta}_{2f}, \tau^e_{2l}, \tilde{\tau}^e_{2f})$,
$\tau_{1f}(t)=\tau_{1f}(\tilde{\theta}_{1l}, \theta_{1f}, \tilde{\tau}^e_{1l}, \tau^e_{1f})$, and
$\tau_{2f}(t)=\tau_{2f}(\tilde{\theta}_{2l}, \theta_{2f}, \tilde{\tau}^e_{2l}, \tau^e_{2f})$,

{\bf Definition 1: Perfectly undetectable attacks on bilateral teleoperation systems from follower and leader perspectives:}
An attack is perfectly undetectable when: 1) The perceived dynamics of the follower manipulator by the leader are identical to those without the attack, AND 2) The perceived dynamics of the leader manipulator by the follower are identical to those without the attack.

{\bf Corollary 1: Perfectly undetectable reflection attack about the yaw axis.}
Given the perfectly undetectable attacks from the controller's perspective in Section \ref{perfectFDIAsinglemanipulator}, its extension to a bilateral control system is given as follows:
\begin{eqnarray}
        \bm{S}_l &=& {\rm diag} (-1, 1, -1 , 1), \\
    \bm{S}_f &=& {\rm diag} (-1, 1, -1 , 1), \\
    \bm{d}_l &=& [2 \theta_{1f}(0), 0, 0, 0]^T, \\
    \bm{d}_f &=& [2 \theta_{1l}(0), 0, 0, 0]^T.
\end{eqnarray}

\section{Experiments}
\label{experiments}

\subsection{Experimental teleoperation system}
A teleoperation system developed in \cite{TakanashiMS} was implemented between Georgia Tech in Atlanta and the University of Electro-Communications (UEC) in Tokyo. 
The leader manipulator, located in Atlanta, U.S., is operated by a human, while the follower manipulator is situated in Tokyo, Japan. 
Each device is a two-axis manipulator equipped with AC servo motors controlling the yaw and pitch axes. 
The system uses Linux CentOS 8.3 and the Advanced Robot Control System (ARCS) V6 for real-time encrypted control. See Table \ref{tab:spec} for detailed specifications. Communication between the nodes is facilitated via a LAN using UDP socket communication through Georgia Tech's GlobalProtect VPN system, enabling secure, real-time data exchange that currently reports $<$10ms communication latency between Atlanta and Tokyo in an asynchronous mode. 

\begin{figure}
    \centering
    \includegraphics[width=1.0\linewidth]{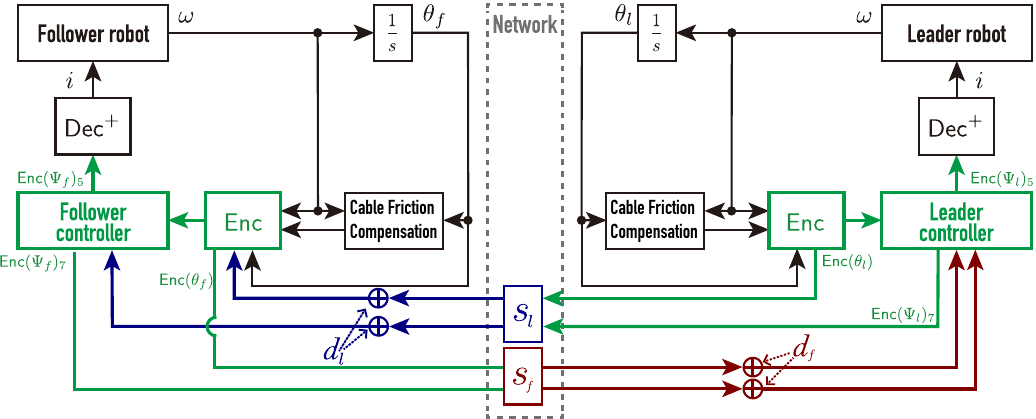}
    \caption{Block diagram of the encrypted four-channel bilateral controller.}
   \label{fig:bd_enc_4ch}
\end{figure}

The controller consists of observers for reaction force estimation and PD and P controllers for position and force control. 
To compute the control commands of the follower, the leader sends the encrypted angles $\theta_{1l}$, $\theta_{2l}$ and estimated refection forces $\hat\tau^e_{1l}$, $\hat\tau^e_{2l}$ to the follower over the network.
The controller encryption~\cite{Kogiso15} with the ElGamal cryptosystem~\cite{ElGamal85} is applied to conceal signals and parameters.
Fig. \ref{fig:bd_enc_4ch} indicates the block diagram of the encrypted controller.
Modules resembling FDIA  by the attacker were added as shown in Fig. \ref{fig:bd_enc_4ch}. In this paper, FDIA has been implemented within the same control system for simplicity, rather than introducing a physically separated 3rd party attack module, such as the one presented in the authors' past paper \cite{Kosugi2024}.

\begin{table}[tb]
\caption{Bilateral control system specifications.}
\label{tab:spec}
\centering
\begin{tabular}{ll}
\hline
\textbf{Yaw axis motor} & MITSUBISHI HK-KT-43W \\
Servo amplifier & MITSUBISHI MR-J5-40A \\
Rated power & {400}{W} \\
Rated torque &  {1.3}{Nm} \\
Rated Current &  {2.6}{A} \\
\hline
\textbf{Pitch axis motor} & MITSUBISHI HK-KT-7M3W \\
Servo amplifier & MITSUBISHI MR-J5-70A \\
Rated power &  {750}{W} \\
Rated torque &  {2.4}{Nm} \\
Rated Current &  {4.7}{A} \\
\hline
\textbf{PC} & 
CPU Intel Core i7-10700K ({5.1}{GHz}) \\
Memory &  {16}{GB} \\
OS & CentOS 8.3 \\
D/A Board & Interface PEX-340216 ( {16}{bit}) \\
Counterboard & Interface PEX-632104 ( {32}{bit}) \\
\hline
\end{tabular}
\end{table}

\subsection{Implementation of encrypted control system and its malleability}

Multiplication of ciphertext control commands is used to keep signals and parameters encrypted.
This paper uses ElGamal encryption~\cite{Teranishi20_1} to encrypt controller signals and parameters, following the encrypted controllers manner~\cite{Kogiso15}.
The cryptosystem consists of three algorithms $(\mathsf{Gen},\mathsf{Enc},\mathsf{Dec})$, where
%
$\Gen:\S\to\K=\Kp\times\Ks$, 
$\lambda\mapsto(\pk,\sk)=((\G,q,g,h),s)$;
$\Enc:\M\times\Kp\to\C$, 
$(m,\pk)\mapsto c=(c_1,c_2)=(g^{r}\bmod p,mh^{r}\bmod p)$;
$\Dec:\C\times\Ks\to\M$, $((c_{1},c_{2}),\sk)\mapsto{c_{1}}^{-s}c_{2}\bmod p$;
$\Gen$ is a key-generation algorithm, 
$\Enc$ is an encryption algorithm, 
$\Dec$ is a decryption algorithm, 
$\pk$ is a public key, 
$\sk$ is a secret key, 
$\lambda$ is a security parameter, 
$q$ is a $\lambda$-bit prime, and 
$p=2q+1$ is a safe prime.
Parameter $g$ represents a generator of a cyclic group $\G\coloneqq\{g^{i}\bmod p\mid i\in\Z_{q}\}$ such that $g^{q}\bmod p=1$, $\Z_q:=\{z\in\Z \mid 0\leq z\leq q\}$, $h=g^{s}\bmod p$, $\M=\G$ and $\C=\G^{2}$.
$r$ and $s$ are random numbers in $\Z_{q}$. 

The encryption scheme allows multiplication of plaintext through operations on ciphertext, which is called multiplicative homomorphism.: 
\begin{align*}
    &\Dec\left(\Enc(m_1,\pk \otimes \Enc(m_2,\pk)\bmod p,\sk
    )\right)\\
    &=m_1m_2,\,\forall k\in\Z_{\geq 0}:=\{z\in\Z\mid z \geq 0\},
\end{align*}
where $\otimes$ denotes an elemental-wise product, $m_1,m_2\in \M$ are messages, $\pk$ and $\sk$ are the public and secret keys, respectively.


Although homomorphic encryption (HE) allows for secure computation without revealing vital information such as control gains or observables, HE is inherently vulnerable to malleability-based FDIA.
Malleability is defined as a property of a specific class of homomorphic encryption methods that allows arithmetic operations on ciphertext without knowing encryption keys.
FDIA based on malleability is defined as follows: For any $c=\Enc(m,\pk)\in\C$,
\begin{equation}
\label{mallebility}
c'=(c'_1,c'_2):=(c_1,kc_2 \;{\rm mod} \; p)
\end{equation}
where $k$ is an attack parameter and $c'$ is alternated cipher.
Decryption $\Dec(c',\sk)$ generates $km$ if $km\in\M$.
Multiplication of $c_{2}$ by $k$ results in manipulating a corresponding plaintext $m$, i.e., $\text{Dec}(c',\sk)=k m$.


\subsection{Attack scenarios and experimental procedure}
\label{scenario}
For simplicity, zero initial conditions are assumed. Also, assuming malleability works with all $k$ in real numbers, three attack scenarios were considered for the four-channel bilateral control system in Fig.~\ref{fig:bd_enc_4ch}. The normal operation corresponds to no attacks on the communication line.
The reflection and scaling attacks were realized by the malleability of encrypted signals (\ref{mallebility}) sent over networks.
The reflection attack was realized as multiplication with the gain $k=-1$ for all encrypted signals on the network.
The scaling attack was realized by introducing a switching gain of $k=2$ and its reciprocal $1/2$ depending on the communication direction.
\begin{itemize}
    \item {\bf Normal operation} (no attack):  \\$\bm{S}_l=\bm{I}_4, \bm{d}_l=\bm{0}, \bm{S}_f=\bm{I}_4,\bm{d}_f=\bm{0}$.
    \item  {\bf Scenario 1 - Reflection attack about the yaw axis}: \\ \
    $\bm{S}_l={\rm diag} (-1, 1, -1 , 1)$,
    $\bm{d}_l=\bm{0}$, 
    $\bm{S}_f={\rm diag} (-1, 1, -1 , 1)$,
    $\bm{d}_f= \bm{0}$.
    \item {\bf Scenario 2 - Scaling attack ($\times$2 of the follower motion)}: $\bm{S}_l={\rm diag} (2, 2, 2 , 2), 
    \bm{d}_l=\bm{0}$, $\bm{S}_f={\rm diag} (0.5, 0.5, 0.5 , 0.5)$, $\bm{d}_f=\bm{0}$.
\end{itemize}
 Note that scaling attacks are infeasible for the 2-DOF nonlinear dynamics, and are only applicable when the pitch axis is fixed, making the manipulator a 1-DOF linear system.

The leader was manipulated in a sinusoidal-like pattern along the pitch axis followed by the yaw axis.
Leader and follower were uninhibited in the intended motion path in the first case. In the second control case, a metal block was placed to impede the motion path of the follower. As a result, the leader's motion path would be unable to complete the entire motion path, with the operator feeling the interaction force between the follower and the metal block.

\begin{figure}
    \centering
    \includegraphics[width=1.0\linewidth]{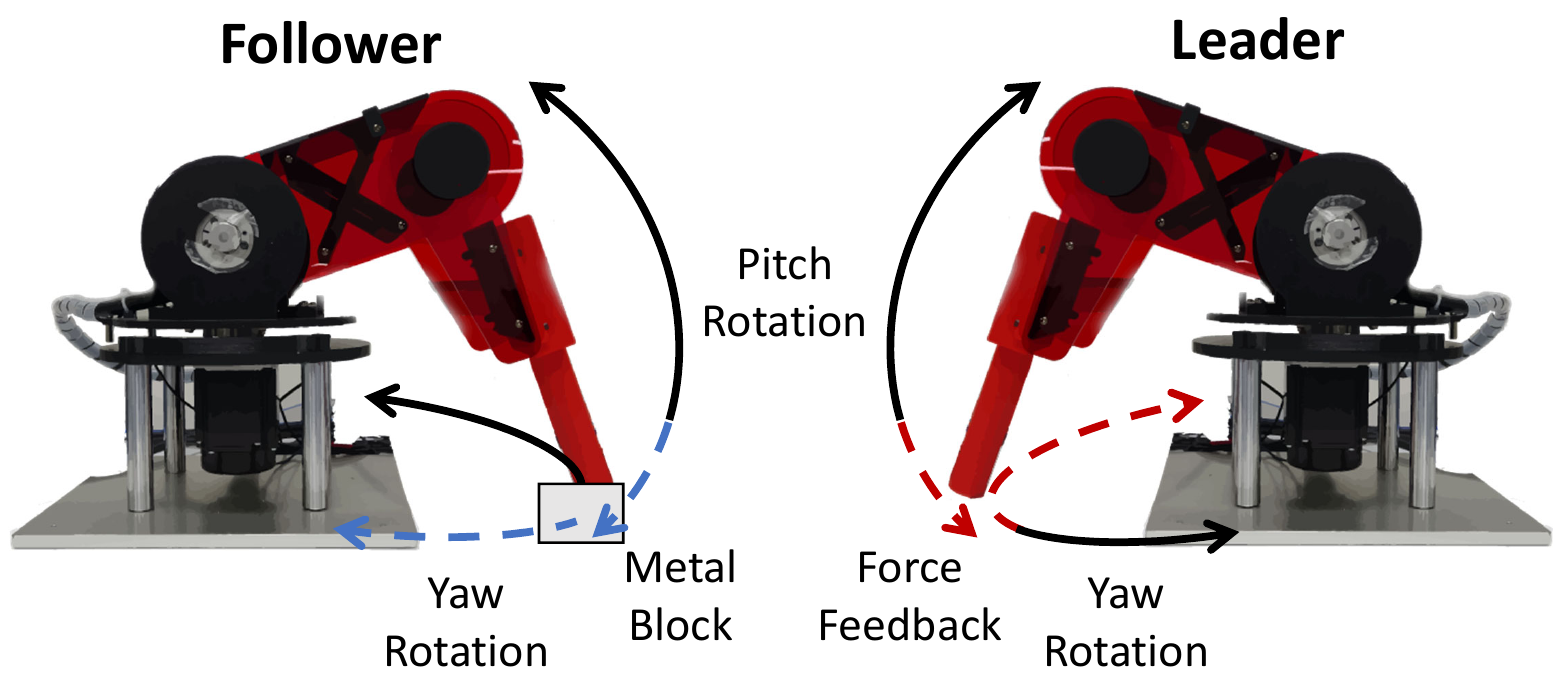}
    \includegraphics[width=.95\columnwidth]{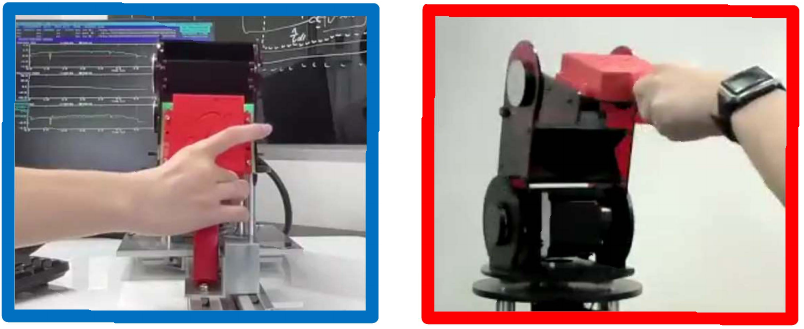}
    \caption{Experimental procedure. The leader is manipulated first in the pitch axis, then the yaw axis. 
    The follower is under the effect of FDIA according to Section \ref{scenario} in each trial.
    The follower motion was uninhibited in Figs. \ref{fig:normal_no_collision} and \ref{fig:enc_negation}. 
    The metal block shown impeded follower arm's motion in Figs. \ref{fig:normal_collision} and \ref{fig:negation_collision}.
    }
    
    \label{fig:Exp Diagram}
    
\end{figure}

\begin{figure}[t]
    \centering
    \subfloat[Rotation Yaw angle: $\theta_{1l}$ and $\theta_{1f}$.]{\includegraphics[width=.45\columnwidth]{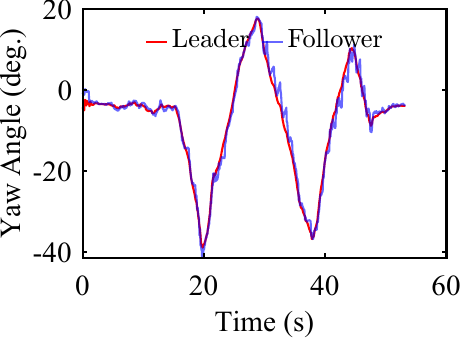}\label{fig:yaw_theta_no_collision}} \quad
    \subfloat[Estimated Yaw reaction force: $\hat\tau^e_{1l}$ and $\hat\tau^e_{1f}$.]{\includegraphics[width=.45\columnwidth]{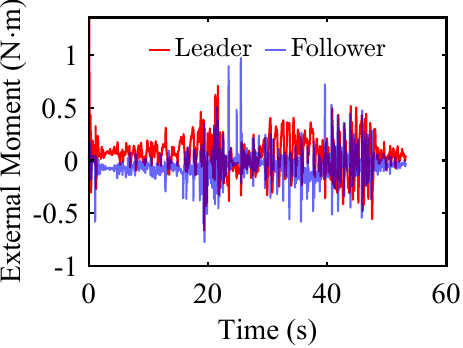}\label{fig:yaw_tau_no_collision}} \\
    \subfloat[Rotation Pitch angle: $\theta_{2l}$ and $\theta_{2f}$.]{\includegraphics[width=.45\columnwidth]{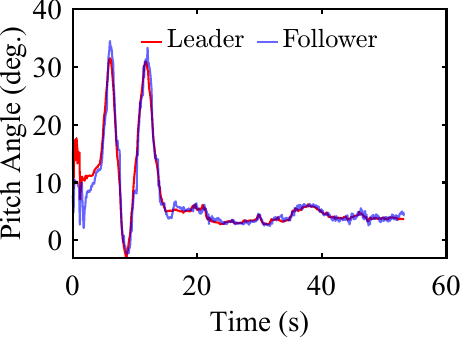}\label{fig:pitch_theta_no_collision}} \quad
    \subfloat[Estimated Pitch reaction force: $\hat\tau^e_{2l}$ and $\hat\tau^e_{2f}$.]{\includegraphics[width=.45\columnwidth]{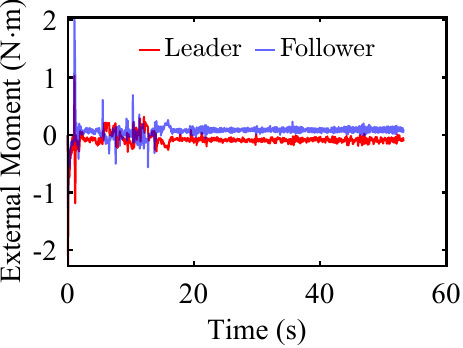}\label{fig:pitch_tau_no_collision}}
    
    \caption{Experimental results of the pose tracking (no attack). 
    }
    \label{fig:normal_no_collision}
\end{figure}

\begin{figure}[t]
    \centering
    \subfloat[Rotation Yaw angle: $\theta_{1l}$ and $\theta_{1f}$.]{\includegraphics[width=.45\columnwidth]{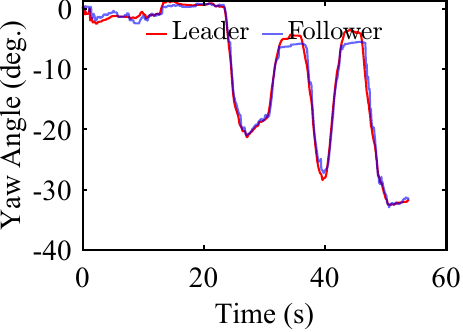}\label{fig:yaw_theta_metal_collision}} \quad
    \subfloat[Estimated Yaw reaction force: $\hat\tau^e_{1l}$ and $\hat\tau^e_{1f}$.]{\includegraphics[width=.45\columnwidth]{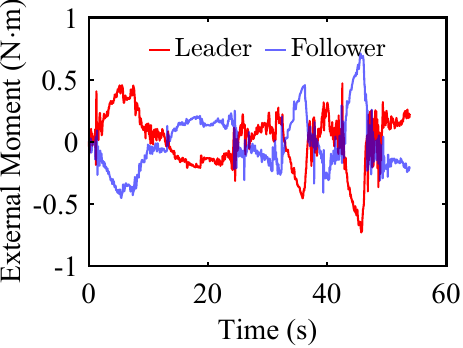}\label{fig:yaw_tau_metal_collision}} \\
    \subfloat[Rotation Pitch angle: $\theta_{2l}$ and $\theta_{2f}$.]{\includegraphics[width=.45\columnwidth]{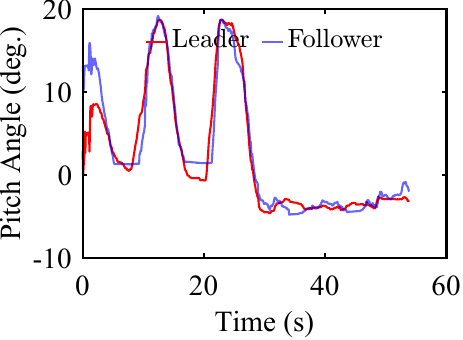}\label{fig:pitch_theta_metal_collision}} \quad
    \subfloat[Estimated Pitch reaction force: $\hat\tau^e_{2l}$ and $\hat\tau^e_{2f}$.]{\includegraphics[width=.45\columnwidth]{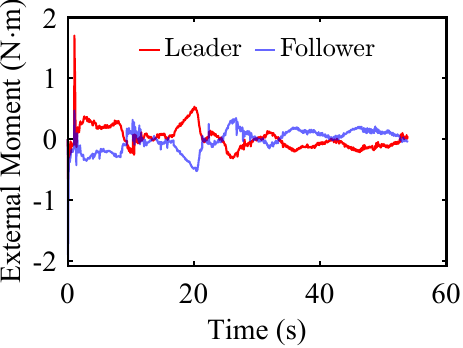}\label{fig:pitch_tau_metal_collision}}
	\caption{Experimental results of the pose tracking (no attack) collided with a metal block.}
	\label{fig:normal_collision}
\end{figure}

\subsection{Results}


All experiments were executed with a sampling time of 20~ms.
The attacks start from $t=0$ to the operation ends.
For simplicity, the initial conditions of each robot is set to be zero at startup.
This reduces the additive element $d_l$ and $d_f$ to the zero vector.

To verify the functionality of the system, two control cases under no FDIA were examined.
The pose tracking for the control case (no collision, no attack) is shown in Fig. \ref{fig:normal_no_collision}, showing close tracking of angular position in both manipulators. In contrast, the control collision case Fig. \ref{fig:normal_collision} shows a flat slope at the follower's angular position at about -5 degrees in the yaw axis and 3 degrees in the pitch axis, indicating where the metal block was in the motion path. As the follower collided with the metal block, the magnitude of the external moment increases for both the follower and the leader (Figs. \ref{fig:normal_collision}(b) and \ref{fig:normal_collision}(d)). The extended motion of the leader is also relatively flattened from when the operator sensed the force feedback caused by the follower's collision at around 5 and 17 seconds for pitch axis, and 30 and 45 seconds for the yaw axis.

\begin{figure}[t]
    \centering
    \subfloat[Rotation Yaw angle: $\theta_{1l}$ and $\theta_{1f}$.]{\includegraphics[width=.45\columnwidth]{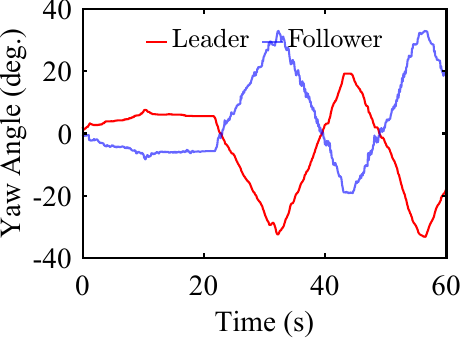}\label{fig:yaw_theta_negation_no_collision}} \quad
    \subfloat[Estimated Yaw reaction force: $\hat\tau^e_{1l}$ and $\hat\tau^e_{1f}$.]{\includegraphics[width=.45\columnwidth]{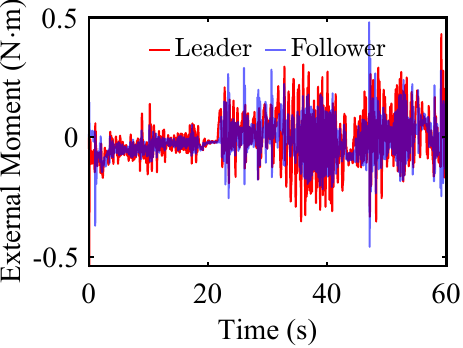}\label{fig:yaw_tau_negation_no_collision}} \\
    \subfloat[Rotation Pitch angle: $\theta_{2l}$ and $\theta_{2f}$.]{\includegraphics[width=.45\columnwidth]{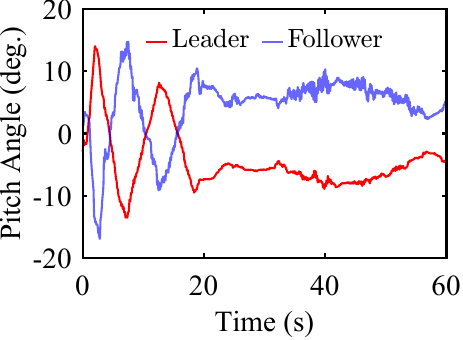}\label{fig:pitch_theta_negation_no_collision}} \quad
    \subfloat[Estimated Pitch reaction force: $\hat\tau^e_{2l}$ and $\hat\tau^e_{2f}$.]{\includegraphics[width=.45\columnwidth]{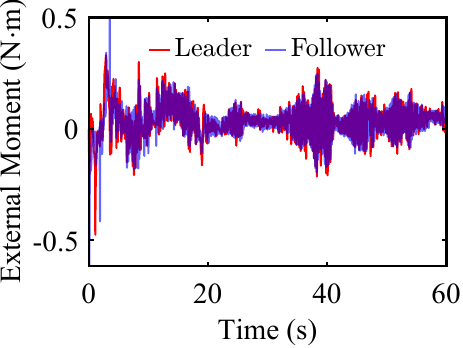}\label{fig:pitch_tau_negation_no_collision}}
	\caption{Experimental results of the pose tracking under a reflection attack to encrypted signals (Scenario 1). The attack made the follower motion to the opposite.}
	\label{fig:enc_negation}
\end{figure}

\begin{figure}[t]
    \centering
    \subfloat[Rotation Yaw angle: $\theta_{1l}$ and $\theta_{1f}$.]{\includegraphics[width=.45\columnwidth]{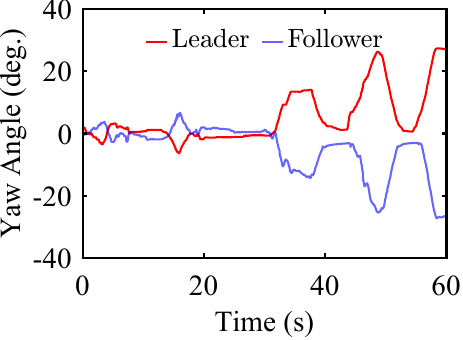}\label{fig:yaw_theta_negation_metal_collision}} \quad
    \subfloat[Estimated Yaw reaction force: $\hat\tau^e_{1l}$ and $\hat\tau^e_{1f}$.]{\includegraphics[width=.45\columnwidth]{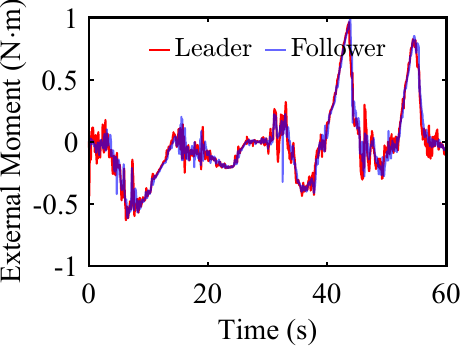}\label{fig:yaw_tau_negation_metal_collision}} \\
    \subfloat[Rotation Pitch angle: $\theta_{2l}$ and $\theta_{2f}$.]{\includegraphics[width=.45\columnwidth]{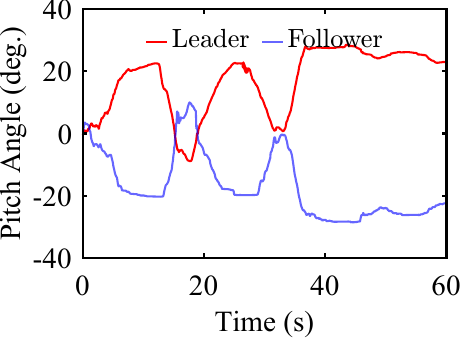}\label{fig:pitch_theta_negation_metal_collision}} \quad
    \subfloat[Estimated Pitch reaction force: $\hat\tau^e_{2l}$ and $\hat\tau^e_{2f}$.]{\includegraphics[width=.45\columnwidth]{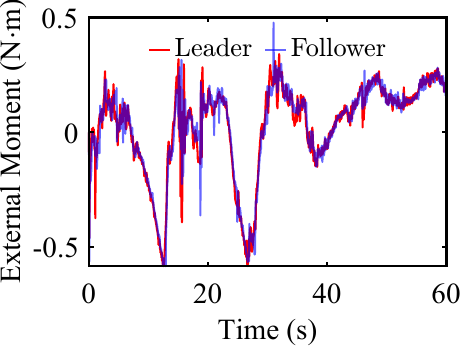}\label{fig:pitch_tau_negation_metal_collision}}
	\caption{Experimental results of the pose tracking under a reflection attack (Scenario 1) when colliding with a metal block.}
\label{fig:negation_collision}
    \centering
    \subfloat[Rotation Yaw angle: $\theta_{1l}$ and $\theta_{1f}$.]{\includegraphics[width=.45\columnwidth]{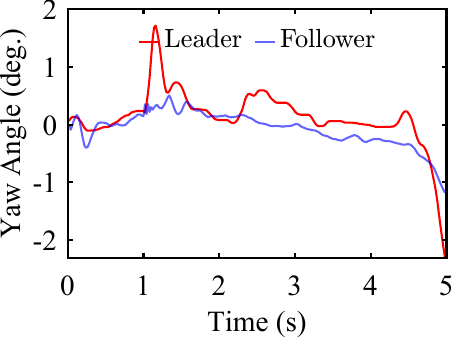}\label{fig:yaw_theta_negation_metal_collision}} \quad
    \subfloat[Estimated Yaw reaction force: $\hat\tau^e_{1l}$ and $\hat\tau^e_{1f}$.]{\includegraphics[width=.45\columnwidth]{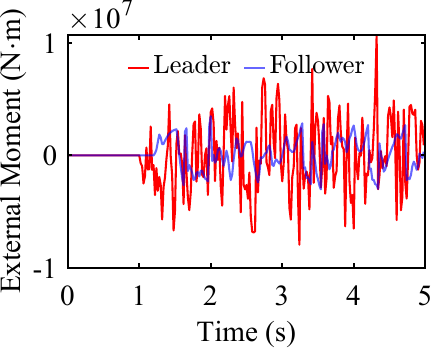}\label{fig:yaw_tau_negation_metal_collision}} \\
    \subfloat[Rotation Pitch angle: $\theta_{2l}$ and $\theta_{2f}$.]{\includegraphics[width=.45\columnwidth]{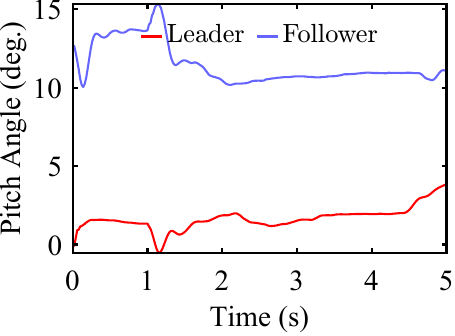}\label{fig:pitch_theta_negation_metal_collision}} \quad
    \subfloat[Estimated Pitch reaction force: $\hat\tau^e_{2l}$ and $\hat\tau^e_{2f}$.]{\includegraphics[width=.45\columnwidth]{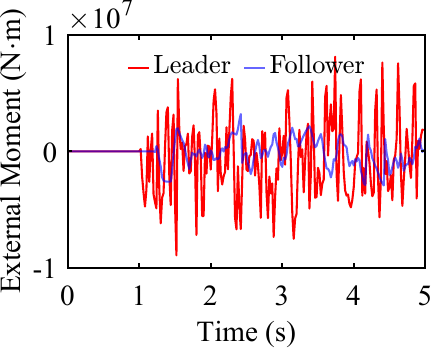}\label{fig:pitch_tau_negation_metal_collision}}
	\caption{Experimental results of pose tracking when under a scaling attack (Scenario 2) as a failure case.}
	\label{fig:failure_case}
\end{figure}

The reflection attack (Scenario 1) was applied to both robots through the malleability attack shown in (\ref{mallebility}). 
Since the ElGamal cipherspace is defined as a cyclic group, the inverse of the original message always exists, which makes this FDIA successful. 
Figs. \ref{fig:enc_negation} and \ref{fig:negation_collision} show the operation of the system under Scenario 1.
The system under the attack behaves in a mirrored manner in which the interaction force was also observed in the opposite direction.
This could lead to dangerous situations if the human operator is not able to detect the attack in time. 

While the existence of perfectly undetectable FDIA has been mathematically demonstrated in the previous section, implementing such an attack requires additional effort from the attacker to overcome practical complications including computational delays and synchronization of onset between observables and command attacks. Note that none of them was significant in the current experiments, but some may be significant in different configurations. 
Strictly speaking, perfectly undetectable FDIAs should be verified by observing identical error dynamics across different yet successful attack scenarios. Unlike \cite{ueda2024affinetransformationbasedperfectlyundetectable}, note that identical leader and follower movements were not reproduced across different trials due to the nature of the human-in-the-loop system.

Fig. \ref{fig:failure_case} shows results when a scaling attack (Scenario 2) was applied. 
The choice of attack parameters $\varphi_u = [2,2]$ is not a valid automorphism, as scaling of a trigonometric function disrespects the group action of angle addition.
This implies the dynamics of the manipulator inhibits the scaling attack.
In addition to the limitation in dynamics, the ciphersystem also limits the range of applicable FDIA.

The erroneous values shown in Fig. \ref{fig:failure_case} (b) and (d) result from failure to decrypt following the FDIA.
Since the ElGamal cipherspace is defined over the integers, dividing an odd ciphertext value by 2 disrespects the group structure.
In this case the message is no longer an element in the cipherspace.
The malleability attack fails and the original message cannot be recovered anymore.
In this case, values far beyond normal operation indicate that an attack has been applied to the system, making the attack detectable.

\section{Conclusion}

This paper demonstrated that robot manipulator dynamics with inherent symmetry are susceptible to a specific type of perfectly undetectable FDIA. Using a typical 2-axis manipulator dynamic model as a representative system, the paper showed that the trigonometric functions present in the dynamic equations are particularly vulnerable to a reflection-type FDIA. This vulnerability arises from the symmetric properties of these functions, which can be used by attacker to alter the robot's behavior while remaining undetected. As a case study, a bilateral teleoperation system was investigated as to how such perfectly undetectable reflection attacks may be implemented. Even if signals in the communication channels are encrypted, the attack can still be realized via the malleability of homomorphic encryption. Future work includes investigations of attack synchronization, communication delays as well as countermeasures to prevent malleability-based attacks. 

\appendices

 \section{Perfectly undetectable FDIA from the plant's perspective}
\label{appen_perfectFDIAplant}

{\bf Definition A1: Perfectly undetectable FDIA from the plant's perspective.} (Milosevic 2021 \cite{Sandberg22,GRACY21}). Let $y(x(0),u,a)$ denote the response of the system for the initial condition $x(0)$, input $u(t)$, and attack signal $a(t)$. The attack is perfectly undetectable if 
\begin{equation}
    y(x(0),u,a)=y(x(0),u,0), t \geq 0.
    \label{perfectFDIAplantdef}
\end{equation}

The attacker does not leave any traces in the measurements of $y$, and can impact the system’s performance or behavior without being noticed by an attack detector that utilizes $y$ for attack detection. Research showed that (\ref{perfectFDIAplantdef}) can be achieved by zero dynamics attacks with the existence of transmission zeros \cite{Sandberg22,Milošević20,Mao20}. 
In that definition, the detector receives ground truth observables without being compromised, which is not assumed in this particular paper. 

\bibliographystyle{IEEEtran}
\bibliography{citations}
\end{document}